\documentclass{article}

\usepackage{arxiv}

\usepackage[utf8]{inputenc} 
\usepackage[T1]{fontenc}    
\usepackage{hyperref}       
\usepackage{url}            
\usepackage{booktabs}       
\usepackage{amsfonts}       
\usepackage{nicefrac}       
\usepackage{microtype}      
\usepackage{lipsum}		
\usepackage{graphicx}
\usepackage{doi}
\usepackage{multirow}
\usepackage[dvipsnames]{xcolor}

\title{\emph{Rendezvous in Time}: An Attention-based Temporal Fusion approach for Surgical Triplet Recognition}

\author{ 
        {
        \hspace{12mm}\href{https://orcid.org/0000-0002-6021-6132}{\includegraphics[scale=0.06]{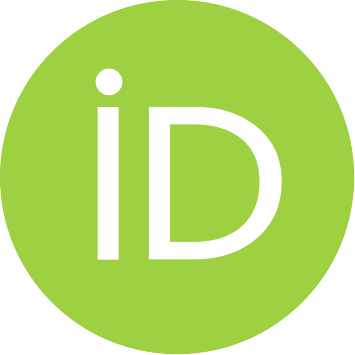}} Saurav Sharma} \\
	{\hspace{14mm}ICube Laboratory}\\
	{\hspace{10mm}University of Strasbourg, France}\\
	{\hspace{14mm}\texttt{ssharma@unistra.fr}} \\
	\And
	{
        \hspace{23mm}\href{https://orcid.org/0000-0003-4777-0857}{\includegraphics[scale=0.06]{orcid.pdf}} Chinedu Innocent Nwoye} \\
	{\hspace{22mm}ICube Laboratory}\\
	{\hspace{22mm}University of Strasbourg, France}\\
	{\hspace{22mm}\texttt{nwoye@unistra.fr}}\\
	\And
        \And
 	{
         \hspace{1mm}\href{https://orcid.org/0000-0002-7559-3328}{\includegraphics[scale=0.06]{orcid.pdf}} Didier Mutter} \\
        IHU Strasbourg, France\\
        University Hospital of Strasbourg, France\\
	\texttt{didier.mutter@ihu-strasbourg.eu} \\
        \And
	{
        \hspace{1mm}\href{https://orcid.org/0000-0002-5010-4137}{\includegraphics[scale=0.06]{orcid.pdf}} Nicolas Padoy} \\
        IHU Strasbourg, France\\
        ICube, University of Strasbourg, CNRS, France\\
	\texttt{npadoy@unistra.fr} \\
}

\renewcommand{\headeright}{}
\renewcommand{\undertitle}{}
\renewcommand{\shorttitle}{Rendezvous in Time}

\hypersetup{
pdftitle={A template for the arxiv style},
pdfsubject={q-bio.NC, q-bio.QM},
pdfauthor={David S.~Hippocampus, Elias D.~Striatum},
pdfkeywords={First keyword, Second keyword, More},
}

\date{}

\begin{document}
\maketitle

\begin{abstract}
One of the recent advances in surgical AI is the recognition of surgical activities as triplets of \textlangle{instrument, verb, target}\textrangle{}. Albeit providing detailed information for computer-assisted intervention, current triplet recognition approaches rely only on single frame features. Exploiting the temporal cues from earlier frames would improve the recognition of surgical action triplets from videos.
In this paper, we propose \emph{Rendezvous in Time (RiT)} - a deep learning model that extends the state-of-the-art model, Rendezvous, with temporal modeling.
Focusing more on the verbs, our RiT explores the connectedness of current and past frames to learn temporal attention-based features for enhanced triplet recognition.
We validate our proposal on the challenging surgical triplet dataset, CholecT45, demonstrating an improved recognition of the verb and triplet along with other interactions involving the verb such as \textlangle{instrument, verb}\textrangle{}. Qualitative results show that the RiT produces smoother predictions for most triplet instances than the state-of-the-arts.
We present a novel attention-based approach that leverages the temporal fusion of video frames to model the evolution of surgical actions and exploit their benefits for surgical triplet recognition.
\end{abstract}


\section{Introduction}
Laparoscopy, despite its many advantages, added some level of complexity to surgical procedures \cite{cai4cai}. This brings the need to support surgeon's efforts with highly adaptive AI systems that can learn from streams of digital data~\cite{surgicalds} available in the operating room (OR) to provide intraoperative context-awareness~\cite{cai4cai}.
Bulk of the existing research focuses on surgical phase recognition~\cite{endonet} and instrument spatial and presence detection~\cite{jin2018tool}.
The phase information is coarse-grained, concealing fine details of the activities taking place~\cite{nwoye2021deep}, whereas the instrument bounding box location might not necessarily contain information about interaction with the anatomy.
Even the finer action recognition~\cite{wagner2021comparative} that summarizes scene activity by a single verb, is devoid of instrument and tissue details.

Taking everything into consideration, surgical action triplet is proposed~\cite{katic2014knowledge,tripnet} to formalize and recognize surgical activities as triplets of \textlangle{}{\textit{instrument, verb, target}}\textrangle{} enabling detailed and holistic surgical activity understanding for more helpful AI assistance in the OR. At this milestone, current methods are yet to explore the freely available temporal data in surgical videos. Tripnet~\cite{tripnet}, the first deep learning model on this task, uses a multi-task learning framework to capture instances of instrument, verb, and target from a single image frame and associate them to form a triplet using a 3D interaction function. The state-of-the-art method, Rendezvous (RDV)~\cite{rdv}, improves on the performance using a hybrid attention mechanism, yet on single frames. This could explain why the RDV performs low on some triplets such as \textlangle{}{\textit{irrigator, aspirate, fluid}}\textrangle{}, \textlangle{}{\textit{grasper, retract, omentum}}\textrangle{}, \textlangle{}{\textit{bipolar, coagulate, cystic-pedicle}}\textrangle{}, which are heavily differentiated by temporal dynamics of their instruments and targets. 
Temporal modeling has been employed in many other surgical workflow analyses, e.g., phase recognition~\cite{tecno,mtrcnetcl}, tool tracking~\cite{lstmtool}, to improve performance leveraging inherent motion characteristics in video-based data.

Motivated by the need for temporal modeling in fine-grained analysis, we propose \textbf{RiT}, which stands for \textbf{Rendezvous in Time}, that extends the RDV with temporal modeling.
We focus our temporal modeling on the \textit{verb} component of the triplet following our strong assumption that the verb is the most temporal dependent component of the triplet.
As an example, the triplet \textlangle{clipper, clip, cystic-duct}\textrangle{} would require more than one frame to discern the \textit{clip} activity whereas the \textit{clipper} and \textit{cystic-duct} terms can be obvious from a single frame.
Predicting the verbs from a single frame, as has been the norm in existing models, would be misleading and unreliable, especially in the case of the instrument's occlusion and other procedural noise.
Therefore, the RiT is designed to study the effects of integrating the discriminative temporal context of the verb features into the triplet recognition task. 
We specifically ask - \textit{How can we model the verb features across time with higher precision without sacrificing the triplet recognition accuracy}? 
To answer this research question, RiT employs a novel Temporal Attention Module (TAM) on the verb processing layer of the RDV to aggregate verb features from current and past frames in a late fusion setup. This allows for the learning of local motion representations of instrument-target interactions, which are pivotal for activity understanding.
We further evaluate RiT on the challenging CholecT45 dataset \cite{rdv} and show improvements on verb and triplet recognition as well as on other sub-level associations.

The contributions in this work are two folds: (1) we propose a novel temporal attention mechanism (TAM) targeting the verb component of triplet utilizing current and past frames; (2) we justify our choice of temporal model by providing extensive ablation studies for different configurations of TAM.

\section{Related Works}
\paragraph{\textbf{Surgical Workflow Analysis:}}
Activity recognition is one of the numerous branches of surgical workflow analysis that has deepened in research over time: from coarse-grained phase recognition \cite{dergaseg,funkessltempo,transvnet} to fine-grained gesture \cite{dipietrokinematics}, step \cite{sanatmtl}, and action \cite{wagner2021comparative} recognition.
The finer actions describe surgical activities leaving out details about the instruments and tissues involved in the interactions~\cite{nwoye2021deep}.

\paragraph{\textbf{Surgical Action Triplets:}} 
Surgical triplets in a cholecystectomy procedure present a way to perform fine-grained analysis involving all concerned components by modeling activity as a triplet \textlangle{\textit{instrument, verb, target}}\textrangle{}. 
A 40-video surgical dataset, known as CholecT40 \cite{tripnet}, provides triplet-labeled annotations for research, which is extended to CholecT50 \cite{rdv} with additional 10 videos. 
A related dataset, SARAS-ESAD \cite{bawa2021saras} that formalized activity as only \textlangle{\textit{verb, anatomy}}\textrangle{}, provides bounding box annotations for the surgical instruments on $4$ videos of radical prostatectomy procedures. 
QDNet \cite{lin2022instrument} maintains the complete triplet formalism and provides bounding boxes for both the instruments and targets for cataract surgery.
In terms of method, Tripnet~ \cite{tripnet} utilizes an instrument-centric multi-task model built on a weakly supervised instrument localization module, a class activation-based verb and target recognition, and a learnable interaction matrix to capture triplets. A subsequent approach, RDV~\cite{rdv}, leverages a Transformer-inspired attention method. 
More methods are presented in the 2021 CholecTriplet challenge \cite{ct50challenge2021} for triplet recognition.

\paragraph{\textbf{Temporal Modeling:}}
Modeling temporal dynamics of surgical activities across multiple frames has become a promising way of improving model performance.
Temporal convolution network is applied in TeCNO \cite{tecno} in a stage-wise manner to recognize phases. Trans-SVNet \cite{jin2022trans} utilizes a Transformer to aggregate both spatial and temporal embeddings to predict phase labels. 
ConvLSTM \cite{lstmtool} and MTRCNet-CL \cite{mtrcnetcl} apply recurrent models for instrument motion tracking and phase recognition.
Notwithstanding, temporal modeling is yet to be fully explored on the task of surgical action triplet recognition. Few exceptions in the MICCAI CholecTriplet Challenge \cite{ct50challenge2021} added well-known temporal units like LSTM, GRU, ConvLSTM, etc., on the triplet model predictions. 
Our proposed method innovates by proposing a temporal attention module that aggregates verb features from past frames into the current frame for surgical action triplet understanding.

\section{Methodology}
\subsection{Base Architecture}
Our proposed method is based on RDV \cite{rdv} framework, which is an encoder-decoder style of architecture where the encoder learns individual components - \textit{instrument}(\textbf{i}), \textit{verb}(\textbf{v}), \textit{target}(\textbf{t}), and the decoder models the interactions between components to predict \textit{triplet}(\textbf{ivt}). The encoder takes an ImageNet pre-trained ResNet-18 \cite{resnetcvpr16} to extract visual features. Acting on these features, a weakly supervised localization (WSL) module with two-layer convolution and a global pooling operation models the instrument and generates class activation maps (CAM). These instrument CAMs are input to the Class Activation Guided Attention Mechanism (CAGAM) module as contextual features to learn verb and target features with the help of attention. Additionally, the encoder uses a bottleneck layer to extract low-level scene features \textbf{H\_ivt}. As an input, the decoder then takes the scene features and activation maps from triplet components (\textbf{H\_i}, \textbf{H\_v}, \textbf{H\_t}) and applies layers of multi-head attention to capture their interrelations and generates subtle triplet features \textbf{Y\_ivt}. To induce multi-task training, activation maps \textbf{H\_i}, \textbf{H\_v}, \textbf{H\_t} are average pooled to generate corresponding logits \textbf{Y\_i}, \textbf{Y\_v}, \textbf{Y\_t}. RDV, modeled at the frame level, does not consider the temporal cues from the video data, which can improve verb performance and enhance instrument interaction with the verb component of the triplet. 

\begin{figure}[t!]
    \centering
    \includegraphics[width=0.9\linewidth]{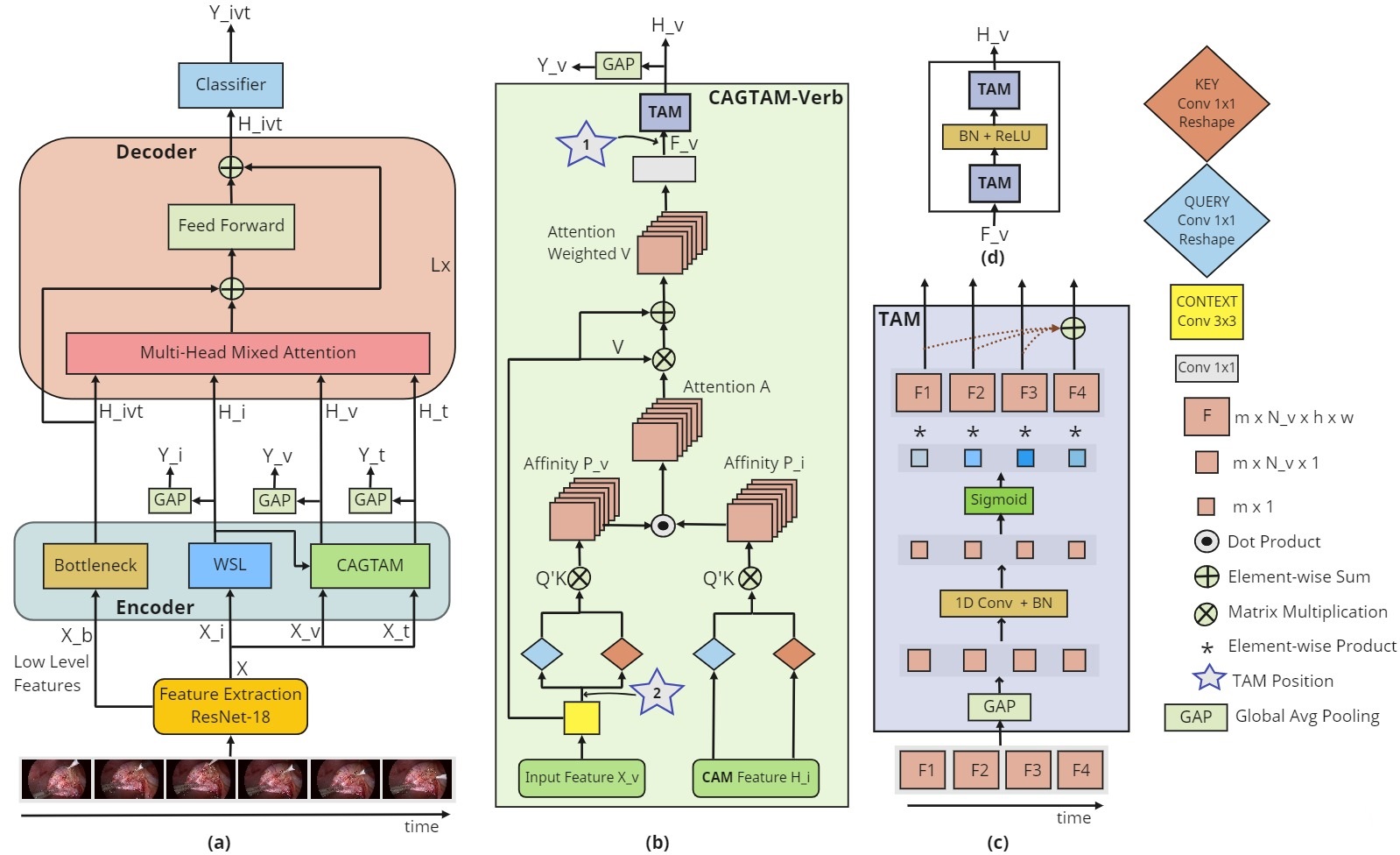}
    \caption{Model Illustration: (a) Rendezvous in Time (RiT) architecture with CAGTAM module extending RDV \cite{rdv}, (b) CAGTAM module layout to process verb features with proposed Temporal Attention Module (TAM), (c) TAM in detail for the temporal fusion of verb features from current and past frames weighted by attention scores following Equation \ref{prop_eq}. (d) Configuration of two TAM layers stacked together. \textbf{Note:} $m$ and N\_v refers to the video clip size and number of verb classes respectively.}
    \label{proposed_method}
\end{figure}

To address the above limitation, we present our proposed framework, \textit{Rendezvous in Time (RiT)}, an RDV-based neural network that adds a new temporal model on the verb component of triplet to capture relations across video frame features. RiT, illustrated in Figure \ref{proposed_method}(a), introduces the Class Activation Guided \textbf{Temporal} Attention Mechanism (CAGTAM) module, adding the long-range capability to the CAGAM module. RiT enhances surgical action triplet recognition by unifying the verb features from past frames to build holistic verb features for the current video frame. 

\subsection{Class Activation Guided Temporal Attention Mechanism (CAGTAM)}
Temporal modeling of verb features is necessary to recognize verb classes such as \textit{clip}, \textit{cut}, and \textit{aspirate} that rely heavily on the temporal dynamics of activity. We begin with the CAGAM module from RDV \cite{rdv} and adapt it to take video clips as input. We then add our novel Temporal Attention Module (TAM) layer at position $1$ denoted by a star in Figure \ref{proposed_method}(b), to fuse past and current frame features. We name this setup consisting of CAGAM with multi-frame clip support and our proposed TAM layer as CAGTAM.

In detail, TAM as shown in Figure \ref{proposed_method}(c),  takes as input a verb feature tensor $\mathcal({F}_-v)_t \in \mathcal{R}^{b \times m \times c \times h \times w}$ at time step $t$, where $c$ denotes total number of verb classes and $m$ the video clip size.
Afterward, TAM applies global average-pooling to get feature $\mathcal({F}_-v)_t  \in \mathcal{R}^{b \times m \times c}$. Then a sequence of \textit{1D-convolution-batch-normalization-sigmoid nonlinearity} generates the probability score $w_{i}$, where $i \in [t-m+1 \dots t]$. Following Equation~\ref{prop_eq}, TAM then fuses verb features from past and current frames, weighted by $w_{i}$ to generate refined verb features $({H}_-v)_t$ at the time step $t$.

\begin{equation}
\label{prop_eq}
({H}_-v)_t = \sum_{i=t-m+1}^{t} w_{i} \odot ({F}_-v[:, i, :, :, :]).
\end{equation}

This enables TAM to model contextual dependencies across past frames while also moderating the noisy or less discriminative features that could generate due to the naive addition of verb features from past frames. Furthermore, due to the utilization of temporal context, TAM learns better instrument-verb semantics which otherwise would be ambiguous in a single-frame context.

\subsection{Verb Feature Fusion}
The position of TAM plays a vital role in deciding when to fuse frame features and directly impacts verb recognition capability. Inspired from \cite{kpvideo}, we explore two strategies to aggregate past and current frame features - early and late fusion marked as position $1$ and $2$ respectively in Figure \ref{proposed_method}(b). In early fusion, TAM processes abstract ResNet-18 features to generate verb features without contextualizing them with instrument CAMs. In late fusion, on the other hand, TAM processes instrument contextualized verb features and exploits useful information present in verb features from past frames to enrich current frame verb features.

\subsection{Video Slicing Input Pipeline}
\label{clip}
To train RiT, we create video clips from $m$ frames following Equation \ref{clip_gen}:

\begin{equation}
\label{clip_gen}
clip_{t} = ([f_{t-m+1} .. f_{t}], ~y_{t}) \; \forall t \in \{0, 1, .., N\},
\end{equation}

\noindent where $N$ is the total number of frames in a video and $y_{t}$ is the ground truth label for the $t^{th}$ frame. To update verb feature at $({F}_-v)_t$, verb features from $[({F}_-v)_{t-m+1}..({F}_-v)_{t-1}]$ are utilized and during training, the loss is computed only at time step $t$. This clip sampling strategy helps to train RiT keeping multiple clips from different videos in a batch and improve the learning process by shuffling across clips and videos. Moreover, the input pipeline is causal as the model learns to predict label $y_{t}$ from clip$_t$ with no access to future frames. 

\section{Experiments}
\subsection{Dataset \& Evaluation Metrics}
We follow the official protocol for data splits and metrics \cite{ctsplits} to conduct our experiments on the multi-label CholecT45 triplet dataset that consists of 6 instrument, 10 verb, 15 target, and 100 triplet classes. The video frames, sampled at a 1fps frame rate, are annotated with binary presence labels of instrument, verb, target, and triplet. We use the official 5-fold cross-validation splits for model evaluation and RDV splits for ablation studies. Also, the logits for instrument, verb, and target components are extracted from main triplet logits as suggested in \cite{ctsplits}. We use video-specific average precision (AP) to evaluate model performance.

\subsection{Implementation Details}
Following \cite{ctsplits}, we resize the video clip frames to $256 \times 448$ resolution and apply data augmentation strategies such as horizontal flipping, and brightness/contrast shift. We remove scaling and vertical flipping augmentation as these hurt the triplet recognition performance. We use weighted binary cross-entropy loss for the multi-label classification task following: 

\begin{equation}
\label{bce_formula}
L = \sum_{c=1}^{C}\frac{-1}{N}\left( W_cy_clog(\sigma(\overline{y}_c)) + (1-y_c)log(1-\sigma(\overline{y}_c)) \right),
\end{equation}

\noindent where $C$ refers to total number of classes, $y_{c}$ and $\overline{y}_{c}$ respectively denotes correct and predicted labels, $\sigma$ the sigmoid activation function and $W_{c}$ is the weight applied for class balancing following \cite{lstmtool}. We apply loss L on instrument, verb, target, and triplet logits to create the total loss $L_{total}$:

\begin{equation}
\label{loss}
L_{total} = L_{Instrument} + L_{Verb} + L_{Target} + L_{Triplet}.
\end{equation} 

Using our presented input pipeline in Section \ref{clip}, we feed to RiT a batch of random video clips, $\mathcal{V} \in \mathcal{R}^{b \times 3 \times m \times h \times w}$, where m is the video clip size, h=256 is the image height and w=448 is the image width. Given the memory constraint during model training, we consider batches of $16$ video clips. The model is trained end-to-end using Stochastic Gradient Descent with $1e^{-6}$ weight decay and a combination of linear and exponential learning rate schedules. The linear learning rate schedule helps to simulate model warmup where it first learns each triplet component and gradually learns triplet information. RiT, implemented in PyTorch, is trained for $50$ epochs on Nvidia V100 and RTX6000 GPUs. We tune model hyper-parameters on the $5$ validation videos.

\begin{table}[ht]
\centering
    \setlength{\tabcolsep}{15pt}
    \caption{\label{crossval} Triplet Recognition AP (\%) on CholecT45 dataset using official cross-validation splits.}
    \resizebox{\textwidth}{!}{
        \begin{tabular}{@{}lcccccccr@{}}\toprule
            \multirow{2}{*}{Method}&\phantom{abc}&
            \multicolumn{3}{c}{Component detection}&\phantom{abc}&
            \multicolumn{3}{c}{Triplet association}\\ \cmidrule{3-5} \cmidrule{7-9} 
             && $AP_{I}$ & $AP_{V}$ & $AP_{T}$ && $AP_{IV}$ & $AP_{IT}$ & $AP_{IVT}$ \\ \midrule
             Tripnet && \textbf{89.9±1.0} & 59.9±0.9 & 37.4±1.5 && 31.8±4.1 & 27.1±2.8 & 24.4±4.7 \\
             RDV && 89.3±2.1 & 62.0±1.3 & 40.0±1.4 && 34.0±3.3 & 30.8±2.1 & 29.4±2.8 \\
             RiT && 88.6±2.6 & \textbf{64.0±2.5} & \textbf{43.4±1.4} && \textbf{38.3±3.5} & \textbf{36.9±1.0} & \textbf{29.7±2.6} \\
            \bottomrule
        \end{tabular}
        }
\end{table}

\subsection{Results}
\begin{figure}[t!]
    \centering
    \includegraphics[width=0.9\linewidth]{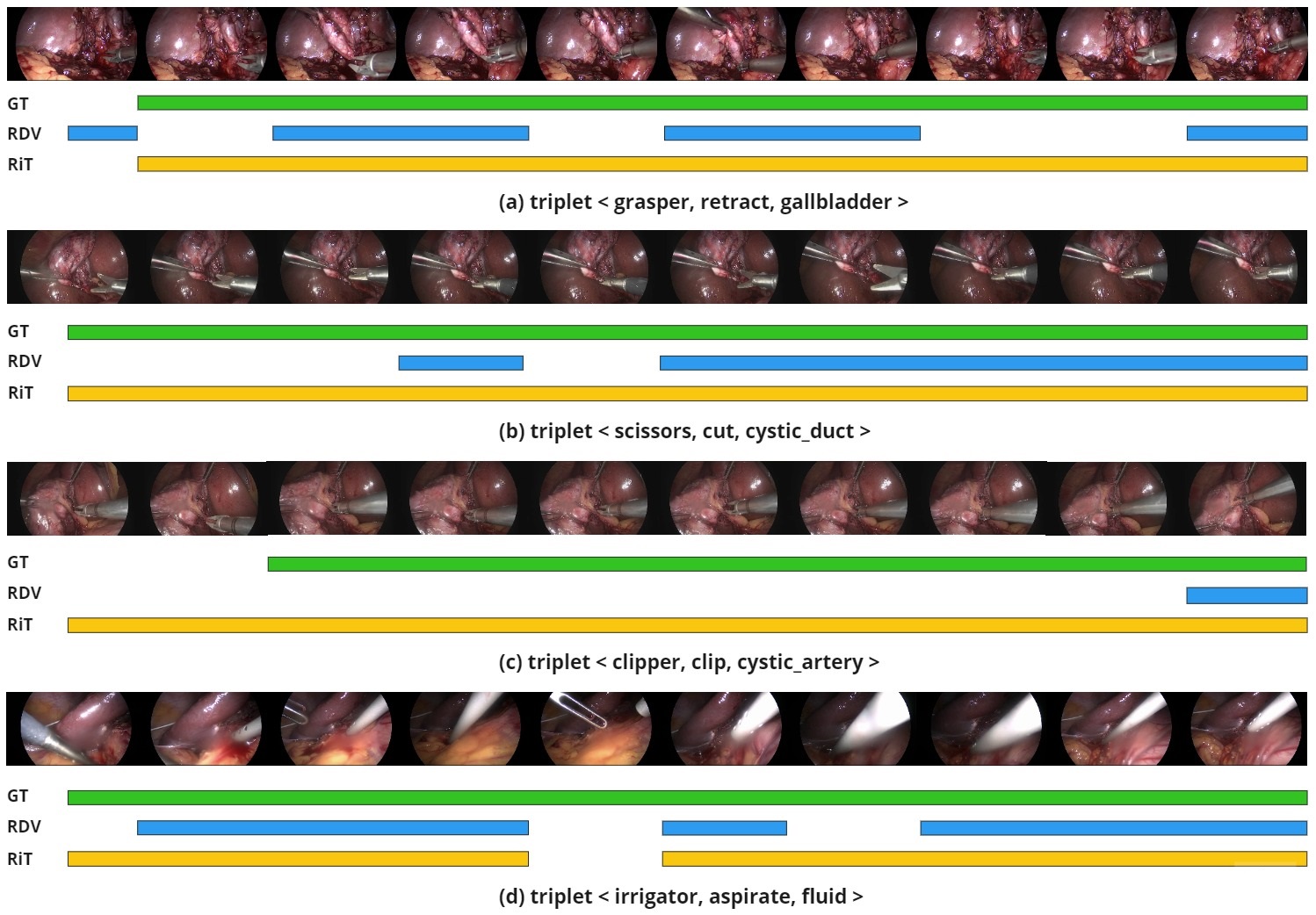}
    \caption{\textbf{Qualitative Results}: Rendezvous in Time (RiT) (\textbf{in orange}) compared with RDV (\textbf{in blue}) and ground truth (GT) (\textbf{in green}) on random images from test videos. For verbs such as \textit{retract}, \textit{cut}, \textit{aspirate}, and \textit{clip}, RiT tries to achieve temporally smooth predictions compared to RDV as seen for triplets. Empty spaces are missed triplet predictions. (Best viewed in color).}
    \label{qual_res}
\end{figure}

We present our results in Table \ref{crossval} for the official 5-fold cross-validation splits where we use RiT with $2$ TAM layers and video clip of $6$ frames. Inside CAGTAM, the $2$ TAM layers are stacked together with BatchNorm and ReLU as shown in Figure \ref{proposed_method}(d). Thanks to the temporal context exploited by the TAM, RiT improves upon state-of-the-art RDV on the verb and triplet recognition further enhancing instrument-verb association. Our novel TAM improves the recognition of $5$ verb classes - \textit{clip}, \textit{cut}, \textit{aspirate}, \textit{pack} and \textit{null-verb} while the performance of other verb classes is similar to that of state-of-the-art. Interestingly, we also observe gains in target recognition and instrument-target association. We believe this is due to the multi-task nature of the training where temporal modeling on verbs impacts target understanding. We further report the classwise performance of verb categories in Table \ref{verb_classwise}. Qualitative results comparing RiT with RDV are shown in Figure~\ref{qual_res}. Note that for baseline results, we consider the PyTorch reproduced version of Tripnet and RDV~\cite{ctsplits}.

\begin{table}[ht]
\centering
    \setlength{\tabcolsep}{45pt}
    \caption{Per-class verb recognition AP (\%) on CholecT45 dataset using official cross validation splits.}
    \label{verb_classwise}
    \resizebox{\textwidth}{!}{
    \begin{tabular}{@{}lcccc@{}}
    \toprule
         Classes && Tripnet & RDV & RiT \\
         \midrule
         grasp     && \textbf{70.5±5.8} & 69.8±3.7 & 69.6±3.9 \\
         retract   && \textbf{90.5±5.4} & 89.7±7.2 & 89.3±9.0 \\
         dissect   && 93.0±2.8 & \textbf{93.2±3.9} & 92.6±2.2 \\
         coagulate && 67.2±6.1 & \textbf{68.7±5.5} & 68.5±6.1 \\   
         clip      && 85.4±6.4 & 85.5±3.7 & \textbf{87.3±5.3} \\
         cut       && 70.5±9.1 & 72.0±4.8 & \textbf{74.9±10.9} \\
         aspirate  && 60.7±9.2 & 57.8±9.9 & \textbf{64.9±7.8} \\
         irrigate  && \textbf{29.6±8.2} & 25.7±5.8 & 22.7±9.0 \\
         pack      && 32.1±9.9 & 31.2±9.9 & \textbf{43.2±16.2} \\
         null-verb && 23.0±2.4 & 24.0±4.1 & \textbf{26.4±4.8} \\
         \midrule
         Mean      && 59.9±0.9 & 62.0±1.3 & \textbf{64.0±2.5} \\
        \bottomrule
    \end{tabular}
    }
\end{table}

\subsection{Ablations}
\paragraph{\textbf{Effect of Position of TAM:}} We investigate the effect of the position of the TAM on triplet recognition performance. As shown in Figure \ref{proposed_method}(b), we add $1$ TAM layer on the verb component at position $1$ and $2$ for late and early fusion respectively. Results in Table \ref{tam_pos} show that late fusion performs better than early fusion due to the verb features contextualized with instrument class activation map adding more discriminative power. Moreover, with both early and late fusion, the performance of the triplet component recognition and their interaction association drops. We keep late fusion (position $1$) setup of TAM for our proposed method.


\begin{table}[ht]
\centering
    \setlength{\tabcolsep}{20pt}
    \caption{\label{tam_pos}Triplet Recognition AP (\%) in early (EF) vs late (LF) fusion.}
    \resizebox{\textwidth}{!}{
        \begin{tabular}{@{}lcccccccr@{}}\toprule
            \multirow{2}{*}{Method}&\phantom{abc}&
            \multicolumn{3}{c}{Component detection}&\phantom{abc}&
            \multicolumn{3}{c}{Triplet association}\\ \cmidrule{3-5} \cmidrule{7-9} 
             && $AP_{I}$ & $AP_{V}$ & $AP_{T}$ && $AP_{IV}$ & $AP_{IT}$ & $AP_{IVT}$ \\ \midrule
             RiT (EF) && 87.3 & 59.3 & 42.1 && 34.9 & 32.5 & 25.1 \\
             RiT (LF) && \textbf{89.5} & \textbf{63.0} & \textbf{45.2} && \textbf{38.7} & \textbf{37.5} & \textbf{30.8} \\
             RiT (EF+LF) && 85.8 & 60.3 & 42.6 && 37.1 & 32.8 & 27.8 \\
            \bottomrule
        \end{tabular}
        }
\end{table}

\paragraph{\textbf{Effect of Clip Size $m$ on Triplet Recognition:}} Ablation study on the clip size $m$ helps in understanding the impact of the duration of temporal context on triplet recognition. Table \ref{clipsize} shows the triplet recognition for different $m$ values where performance diminishes beyond $m=6$. This is because most action triplets are repetitive with short bursts of action for which fusing features from more than the optimal number of frames introduces noise. We keep $m=6$ as the default for our proposed method.

\begin{table}[ht]
\centering
    \setlength{\tabcolsep}{20pt}
    \caption{\label{clipsize}Triplet Recognition AP (\%) on effect of video clip size.}
    \resizebox{\textwidth}{!}{
        \begin{tabular}{@{}lcccccccr@{}}\toprule
            \multirow{2}{*}{Method}&\phantom{abc}&
            \multicolumn{3}{c}{Component detection}&\phantom{abc}&
            \multicolumn{3}{c}{Triplet association}\\ \cmidrule{3-5} \cmidrule{7-9} 
             && $AP_{I}$ & $AP_{V}$ & $AP_{T}$ && $AP_{IV}$ & $AP_{IT}$ & $AP_{IVT}$ \\ \midrule
             \hspace{5mm}4 && 88.5 & 61.9 & 41.9 && 33.2 & 37.3 & 27.9 \\
             \hspace{5mm}6 && \textbf{89.5} & \textbf{63.0} & \textbf{45.2} && \textbf{38.7} & \textbf{37.5} & \textbf{30.8} \\
             \hspace{5mm}8 && 86.5 & 59.0 & 42.0 && 32.8 & 36.2 & 28.7 \\
            \bottomrule
        \end{tabular}
        }
\end{table}

\begin{table}[t!] 
\centering
    \setlength{\tabcolsep}{15pt}
    \caption{\it \label{tamvsconvlstm}Triplet Recognition AP (\%) for TAM vs Other Temporal Models.}
    \resizebox{\textwidth}{!}{
        \begin{tabular}{@{}lcccccccr@{}}\toprule
            \multirow{2}{*}{Method}&\phantom{abc}&
            \multicolumn{3}{c}{Component detection}&\phantom{abc}&
            \multicolumn{3}{c}{Triplet association}\\ \cmidrule{3-5} \cmidrule{7-9} 
             && $AP_{I}$ & $AP_{V}$ & $AP_{T}$ && $AP_{IV}$ & $AP_{IT}$ & $AP_{IVT}$ \\ \midrule
             RiT (with Non-Local Block) && 87.1 & 59.7 & 41.4 && 36.0 & 34.7 & 29.2 \\
             RiT (with ConvLSTM) && 87.8 & 59.5 & 41.1 && 37.1 & 32.8 & 26.6 \\
             RiT (with ConvGRU)  &&\textbf{89.7} & 61.7 & 43.0 && 38.2 & 34.6 & 29.3 \\
             RiT (with TAM) && 89.5 & \textbf{63.0} & \textbf{45.2} && \textbf{38.7} & \textbf{37.5} & \textbf{30.8} \\
            \bottomrule
        \end{tabular}
        }
\end{table}

\paragraph{\textbf{TAM vs Other Temporal Models:}} We also explore other temporal models such as a ConvLSTM, ConvGRU to highlight the impact of TAM. ConvLSTM has been shown to be effective in the temporal modeling of features, particularly for surgical instrument tracking~\cite{lstmtool}. We choose convolutional recurrent unit over standard LSTM/GRU as it maintains structural locality without flattening the features. We put $1$ ConvLSTM layer with 512-dim features at position $1$ like TAM as shown in Figure~\ref{proposed_method}(b). We repeat the same with ConvGRU. Additionally, we also modify $1$ non-local block~\cite{wang2018non} to operate on the spatial verb feature maps. Results in Table~\ref{tamvsconvlstm} demonstrates the power of TAM in modulating the contribution of verb features from past frames over other temporal models.

We provide additional ablation results on TAM in the supplementary material. We also evaluate our model offline in CholecTriplet (MICCAI 2022) Challenge\footnote{\url{https://cholectriplet2022.grand-challenge.org/results}}, where we would rank $4^{th}$ in the triplet recognition leaderboard. This is encouraging as the high-ranked methods employ bounding box annotations to model triplet recognition, whereas RiT relies only on binary presence labels of the triplet and its components. (detailed ranking is in the supplementary material).

\section{Conclusion}
This work introduces a deep learning method for improved surgical action triplet recognition leveraging temporal information in video clips.
We design a Temporal Attention Module (TAM) to modulate verb features from past frames. The learned features are in turn utilized by our final model, RDV in Time (RiT), which extends the frame-based RDV network with attention-based temporal modeling to predict more accurate triplet labels.
Our results on the challenging CholecT45 dataset demonstrate that learning temporally coherent features while focusing attention on the most relevant locations, is crucial to reap benefits from refined verb features. This leads to capturing instrument-tissue interactions across video frames more accurately.

\paragraph{Acknowledgements:}
This work was supported by French state funds managed by the ANR within the National AI Chair program under Grant ANR-20-CHIA-0029-01 (Chair AI4ORSafety) and within the Investments for the future program under Grant ANR-10-IAHU-02 (IHU Strasbourg). It was granted access to the HPC resources of Unistra Mesocentre.

\bibliographystyle{IEEEtran}
\bibliography{references}

\newpage
\renewcommand{\thesubsection}{\Alph{subsection}}

\section*{Additional Results}\label{extra_results}
\begin{table}[ht]
\centering
    \setlength{\tabcolsep}{18pt}
    \caption{\label{tam_ivtcomp}Results AP (\%) on triplet recognition when TAM is added to other triplet components. TAM performs well when added only to the verb component. The verb performance is lowest when TAM is added on verb and target.
}
    \resizebox{\textwidth}{!}{
        \begin{tabular}{@{}lcccccccccccr@{}}\toprule
            \multirow{2}{*}{}&\phantom{a}&&
            \multicolumn{3}{c}{Component detection}&\phantom{a}&
            \multicolumn{3}{c}{Triplet association}\\ \cmidrule{4-6} \cmidrule{8-10} 
             $V$&$I$&$T$& \rule{0pt}{2ex}  $AP_{I}$ & $AP_{V}$ & $AP_{T}$ && $AP_{IV}$ & $AP_{IT}$ & $AP_{IVT}$ \\ \midrule
            \checkmark & &  & \textbf{89.5} & \textbf{63.0} & \textbf{45.2} && \textbf{38.7} & \textbf{37.5} & \textbf{30.8} \\
             \checkmark & & \checkmark & 79.3 & 53.6 & 39.3 && 30.1 & 27.9 & 21.5 \\
             \checkmark & \checkmark && 81.3 & 59.1 & 40.1 && 34.0 & 29.7 & 24.2 \\
             \checkmark & \checkmark & \checkmark & 82.9 & 55.1 & 34.9 && 33.5 & 28.1 & 23.6 \\
            \bottomrule
        \end{tabular}
        }
\end{table}

\begin{table}[ht]
\centering
    \setlength{\tabcolsep}{14pt}
    \caption{\label{chal2022}Results AP (\%) on CholecTriplet2022 Challenge triplet recognition task †. Models marked ${*}$ use temporal information either directly on verb or on the triplet logits. Our method outperforms INTUITIVE CORTEXT ML as their method does not factor instrument features during temporal modeling of verb.}
    \resizebox{\textwidth}{!}{
        \begin{tabular}{@{}lcccccccr@{}}\toprule
            \multirow{2}{*}{Method}&\phantom{abc}&
            \multicolumn{3}{c}{Component detection}&\phantom{abc}&
            \multicolumn{3}{c}{Triplet association}\\ \cmidrule{3-5} \cmidrule{7-9} 
             && $AP_{I}$ & $AP_{V}$ & $AP_{T}$ && $AP_{IV}$ & $AP_{IT}$ & $AP_{IVT}$ \\ \midrule
             SDS-HD              &&  83.78 &	51.98 &	45.91 &&		36.15 &	38.58 &	34.96 \\
             CITI$^{*}$          && 84.04  & 49.53 &	40.25 && 35.33 &	38.95 & 34.53 \\
             2AI-ICVS$^{*}$      && 78.84  & 55.65 & 40.29 && 35.25 & 36.76 & 34.39 \\
             \color{blue} \textbf{RiT}        &&  \color{blue} 80.23 &	\color{blue} 47.73 &	\color{blue} 39.07 &&	\color{blue} 30.88 & \color{blue} 38.71 & \color{blue} 30.94 \\
             WINTEGRAL           &&  80.33 &	50.34 &	38.36 &&		34.92 &	33.01 &	28.99 \\
             CAMMA               &&  78.21 &	46.55 &	35.85 &&		30.89 &	32.80 &	28.90 \\
             URN$^{*}$ && 82.37 & 49.29 & 38.05 && 33.16 & 31.99 & 27.30 \\
             SHUANGCHUN          &&  85.66 &	52.31 &	39.24 &&		34.33 &	32.31 &	27.21 \\
             CAMP                &&  72.03 &	42.31 &	30.40 &&		29.34 &	28.08 &	23.58 \\
             KLIV-IITKGP         &&  66.15 &	39.68 &	32.25 &&		24.54 &	27.07 &	23.07 \\
             SK                  &&  64.91 & 40.39 &	29.92 &&		26.06 &	23.61 &	20.03 \\
             INTUITIVE\_CORTEX\_ML$^{*}$ && 73.57 & 42.11 & 26.27 && 25.88 & 22.80 & 18.84 \\
            \bottomrule
            \multicolumn{9}{l}{$\dag$ \footnotesize{\url{ https://cholectriplet2022.grand-challenge.org/results/}}}\\
        \end{tabular}
        }
\end{table}

\end{document}